%% file: main.tex
\documentclass[a4paper,fleqn]{cas-dc}

\usepackage[authoryear,longnamesfirst]{natbib}
\usepackage[export]{adjustbox}
\usepackage{wrapfig}

\def\tsc#1{\csdef{#1}{\textsc{\lowercase{#1}}\xspace}}
\tsc{WGM}
\tsc{QE}
\tsc{EP}
\tsc{PMS}
\tsc{BEC}
\tsc{DE}
\begin{document}
\let\WriteBookmarks\relax
\def\floatpagepagefraction{1}
\def\textpagefraction{.001}

\shorttitle{agriFrame: Agricultural framework}

\shortauthors{Author 1 et~al.}

\title [mode = title]{agriFrame: Agricultural framework to remotely control a rover inside a greenhouse environment}                      


\author[1]{Saail Narvekar}[type=editor,
                        auid=000,bioid=1,
                        % prefix=Sir,
                        % role=Principal Corresponding Author,
                        orcid=0000-0001-7511-2910]

\cormark[1]


\ead{213316001@iitb.ac.in}

%\ead[url]{www.cvr.cc, cvr@sayahna.org}

\credit{Conceptualization of this study, Methodology, Software}

\affiliation[1]{organization={Department of Computer Science \& Engineering},
    addressline={Indian Institute of Technology Bombay}, 
    city={Mumbai},
    postcode={400076}, 
    country={India}}

\author[1]{Soofiyan Atar}[]

% Credit for the second author
\credit{Data curation, Writing - Original draft preparation}

\author[1]{Vishal Gupta}[%
   % role=Researcher,
   % suffix=Jr,
   ]
\ead{cvr3@sayahna.org}
%\ead[URL]{www.sayahna.org}

% Credit for the third author
\credit{Data curation, Writing - Original draft preparation}

\author[2]{Lohit Penubaku}
%\cormark[2]
%\fnmark[1,3]
\ead{lpenubaku@iitb.ac.in}
% \ead[URL]{www.stmdocs.in}

% Credit for the fourth author
\credit{Data curation, Writing - Original draft preparation}

% Fifth author
\author[1]{Kavi Arya}

% Credit for the fifth author
% \credit{Data curation, Writing - Original draft preparation}

% Address/affiliation for all authors
\affiliation[2]{organization={Department of Electrical Engineering,
    Indian Institute of Technology Bombay},
    addressline={}, 
    city={Mumbai},
    postcode={400076}, 
    country={India}}

\cortext[cor1]{Corresponding author}
%\cortext[cor2]{Principal corresponding author}

\begin{comment}
\fntext[fn1]{This is the first author footnote. but is common to third
  author as well.}
\fntext[fn2]{Another author footnote, this is a very long footnote and
  it should be a really long footnote. But this footnote is not yet
  sufficiently long enough to make two lines of footnote text.}

\nonumnote{This note has no numbers. In this work, we demonstrate $a_b$
  the formation Y\_1 of a new type of polariton on the interface
  between a cuprous oxide slab and a polystyrene micro-sphere placed
  on the slab.
}
\end{comment}

\begin{abstract}
The growing demand for innovation in agriculture is essential for food security worldwide and more implicit in developing countries. With growing demand comes a reduction in rapid development time. Data collection and analysis are essential in agriculture. However, considering a given crop, its cycle comes once a year, and researchers must wait a few months before collecting more data for the given crop. To overcome this hurdle, researchers are venturing into digital twins for agriculture. Toward this effort, we present an agricultural framework(agriFrame). Here, we introduce a simulated greenhouse environment for testing and controlling a robot and remotely controlling/implementing the algorithms in the real-world greenhouse setup. This work showcases the importance/interdependence of network setup, remotely controllable rover, and messaging protocol. The sophisticated yet simple-to-use agriFrame has been optimized for the simulator on minimal laptop/desktop specifications.   

\begin{comment}
    
This research presents a comprehensive framework for enhancing agricultural automation with a focus on tomato harvesting in greenhouse environments. The framework (agriFrame) integrates a novel sim-to-real transfer approach for the entire software stack, ensuring seamless application in real-world scenarios. It features a uniquely designed hardware system that is both smart and adaptable to the complex needs of agricultural environments. Furthermore, agriFrame is equipped with VPN support, enabling remote control and offering operators a 360-degree view of the greenhouse, thus overcoming the limitations of physical presence. At the heart of the agriFrame is a mobile manipulator specifically developed for agricultural tasks, demonstrating a significant advancement in the efficiency and effectiveness of tomato harvesting. This study showcases the integration of cutting-edge technology in agriculture, promising to revolutionize farming practices through improved operational efficiency, remote management capabilities, and innovative hardware design.
\end{comment}
\end{abstract}

% \includegraphics{figs/grabs.pdf}

% Keywords

\maketitle
\input{intro_related}

\input{simulator}

\input{architecture}
\input{results}
\input{acknowledgment}

\section{Bibliography}

\bibliographystyle{cas-model2-names}

\bibliography{main}


\end{document}

%% file: intro_related.tex
\section{Introduction}

Integrating robotics in agriculture holds significant promise for enhancing efficiency and sustainability in farming practices. Mobile robots offer the potential to automate repetitive tasks, optimize crop conditions, and provide real-time insights about crop health, soil conditions, and microclimates. This translates into informed decision-making, increased yields, and reduced operational costs. An intriguing application of agricultural robotics involves the development of autonomous greenhouse systems. These systems can monitor and regulate various environmental factors like temperature, humidity, light, and CO2 levels, creating the ideal crop growth conditions. Through machine learning algorithms, these autonomous greenhouse systems can predict how alterations in environmental variables might impact crop growth and subsequently adjust these conditions. This innovation has the potential to significantly enhance crop yield, quality, and resource efficiency, leading to a reduction in waste and resource consumption.

\begin{figure}[htp]
    \centering
    \includegraphics[width=9cm]{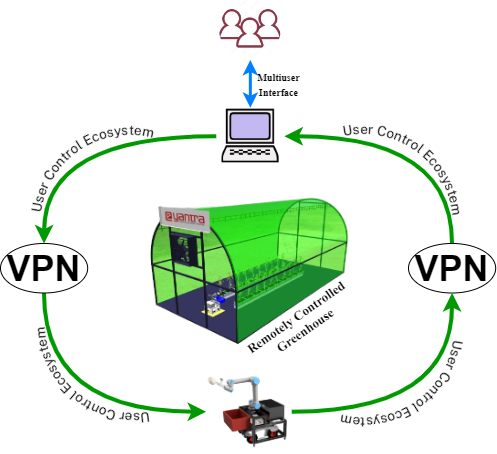}
    \caption{Remote Control Ecosystem}
    \label{fig:galaxy}
\end{figure}

Robotics in agriculture offers an encouraging avenue to improve efficiency, sustainability, and food security. By automating repetitive tasks, optimizing crop conditions, and delivering precise information, robots can aid farmers in augmenting crop yields, cutting down costs, and minimizing the environmental footprint of agricultural practices. Given the escalating global food demand, the incorporation and widespread adoption of robotic systems in agriculture has become imperative to ensure a future that is both sustainable and prosperous.

In a study by Ma et al. (2019) \cite{ma2019greenhouse}, researchers tackled the issue of microclimates affecting plant growth measurements within greenhouses, resulting in data noise. While randomizing pot positions helped mitigate this issue, complete elimination was not achieved. The researchers introduced an approach that optimized the shuffling patterns of plants to mitigate microclimate effects. Their strategy involved creating a computer model that simulated the microclimate conditions within a specific greenhouse. This model incorporated actual design, material, and location information from the Purdue Lily greenhouse in West Lafayette, Indiana. By leveraging simulation results, the researchers optimized the frequency and distance of pot movement. Implementing the new shuffling pattern, informed by the simulation model, led to the removal of more than 90\% of microclimate variance. Notably, this optimized pattern required over 95\% less shuffling effort than non-stop movement, offering a more efficient solution.

Another notable study by Cho et al. (2019) \cite{choab2019review} offers an extensive literature review on greenhouse systems, emphasizing the selection of optimal characteristics for greenhouses across diverse climates and operating conditions. The authors delve into the greenhouse shape and orientation decision-making process based on varying climatic conditions.

Additionally, a simulation tool proposed by Baglivo et al. (2020) \cite{BAGLIVO2020115698} integrates multiple factors influencing the thermal environment within a greenhouse, including external weather conditions, solar radiation, greenhouse design parameters, and crop attributes. By accounting for these variables, the tool can simulate the dynamic behavior of the greenhouse and predict temperature, humidity, and energy requirements for different crop types. The tool's reliability and practical utility are demonstrated through comparisons with real-world data.

The convergence of robotics and agriculture holds immense potential to revolutionize farming practices, enhance productivity, and contribute to a more sustainable future. Pioneering research in this domain addresses challenges related to microclimates, greenhouse optimization, and simulation tools, paving the way for more efficient and effective agricultural practices [\cite{duckett2018agricultural}].

This paper proposes an agricultural framework that can implement greenhouse automation using the algorithms inside a simulator. When the simulations are acceptable, the algorithms are run on a physical rover inside a greenhouse with the capability of controlling the rover remotely. The paper is divided into sections highlighting the core technology  resulting in our agriFrame. Section \ref{agri:frameowrk} 
explains the greenhouse environment in the gazebo simulator, depicting the real world with the model of a rover part of the environment. Section \ref{archi} showcases the agriFrame architecture, which includes rover modification with respect to hardware, setup, and communication in the network between the rover, end user, and rover admin. Section \ref{results} discusses the results/performance of our proposed agriFrame. 

The primary contributions of this research are outlined as follows:
\begin{itemize}
    \item Development of an agricultural framework for deploying and instructing greenhouse rovers (refer to Fig.~\ref{agri:frameowrk}).
    \item Creation of a framework for simulation, development, and remote testing on hardware (Rover).
    \item Introduction of a greenhouse simulator environment to test rovers before real-world deployment.
    \item Establishment of networking capabilities for remote rover control, along with suitable benchmarks.
\end{itemize}

%% file: simulator.tex
\section{AgriFrame Simulator}

A robotics simulator is used to create an application for a physical robot without depending on the physical machine. Simulation software and virtual environments are the two potential tools for accelerating the design and development of agricultural robots \cite{r2018simulation}.

Simulation, in general, refers to the practice of developing and programming virtual models and objects capable of emulating specific tasks, ideas, or a proposal process in the real world. Virtual environments are the replica of the real-world environment in a simulator. The Gazebo simulator was selected as a simulator due to its compatibility with ROS \cite{quigley2009ros} and good community support. ROS, a Robotics Operating System, is not an operating system; rather, it is a framework that provides a structured communications layer above the host operating systems of a heterogenous compute cluster, enabling modularity, tools-based software development and used to built upon by others to build robot software systems which can be useful to a variety of hardware platforms, research settings, and runtime requirements. Gazebo is an open-source 3D robotics simulator. It can use multiple high-performance physics engines, such as ODE, Bullet, etc. (the default is ODE). It provides realistic rendering of environments, including high-quality lighting, shadows, and textures. It can model sensors that "see" the simulated environment. 

\begin{figure*}[!h]
    \centering
    \includegraphics[width =\textwidth]{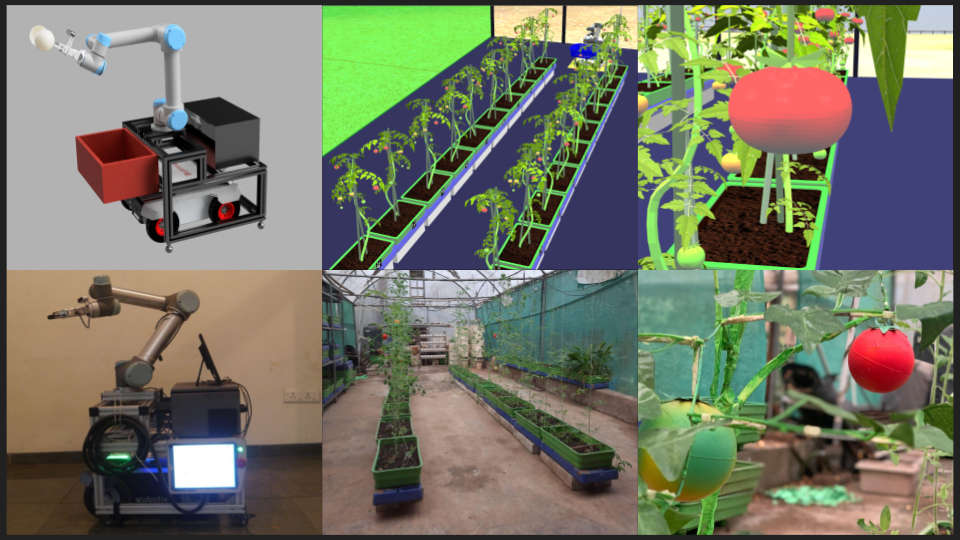}
    \caption{Agribot Sim-to-real}
    \label{fig:agribot-sim-to-real}
\end{figure*}

As per our study, no dynamic agricultural greenhouse virtual environments existed in any simulator. We developed a novel dynamic agricultural greenhouse environment by replicating the actual greenhouse and a robot as shown in Figure 2. Row 1 of Figure 2 shows the digital twin of the robot and greenhouse in simulator. The simulator environment shown in Row 1 column 3 of Figure 2 consists of a pluck-able tomato plant in a greenhouse with tomatoes attached to it. The force/Torque gazebo physics plugin was used to make the pluck-able tomato. A force above the threshold applied to the pedicle of the tomato, it detaches from the plant. The number of plants required can be customized in a greenhouse environment by providing count and position data. In a simulator complex meshes require more computation power to calculate collision physics. Hence, the physics parameters of the greenhouse world environment and robot were optimized by simplifying the mesh files, using simple geometric shapes like spheres, cylinders, and cuboids wherever possible.

The simulator robot model consists of a mobile robot consisting of an industry-standard UR5 robotic arm mounted on a mobile base in a simulated environment based on our physical hardware robot. It consisted of sensors like laser range finders and depth camera sensors in the simulated environment. The mobile base robot was based on a skid steer drive mechanism, gazebo physics plugin was used for motion based on the mechanism. An adaptive gripper was attached as the end effector of the robotic arm. The fingers of the adaptive gripper were customized with a hemisphere-shaped structure for plucking tomatoes. The physics parameters, including the inertial and mass of the robot, were optimized to closely match the actual physical hardware specifications.

%% file: architecture.tex
\section{Architecture}
\label{archi}
The architecture of the agricultural framework consists of 3 blocks: Internet Client, Intranet client, and Hardware setup. Figure 3 shows the architecture diagram. All three blocks are connected to each other through Virtual Private Network(VPN)\cite{ferguson1998vpn}. An Internet Client is a remote client which can access the hardware through VPN. It is present at a remote location from the hardware and can be present anywhere in the world connected to the internet. It consists of  ROS, a  set of software libraries for robot applications. Software in ROS is organized in packages. ROS packages for navigation, perception, and manipulation are present in the Internet Client. RViz package is a 3D visualization tool for ROS to visualize sensor data, robot position, and orientation. MoveIt is a ROS motion planning framework. The remote Client also has live video feedback of a greenhouse environment through RTSP\cite{schulzrinne1998rfc2326} protocol on a browser. It has restricted access to the file system of an Nvidia Jetson TX2 computing board from a Hardware Setup block via Virtual Network Computing(VNC)\cite{richardson1998virtual} protocol. 

\begin{figure*}
    \centering
    \includegraphics[width=\textwidth]{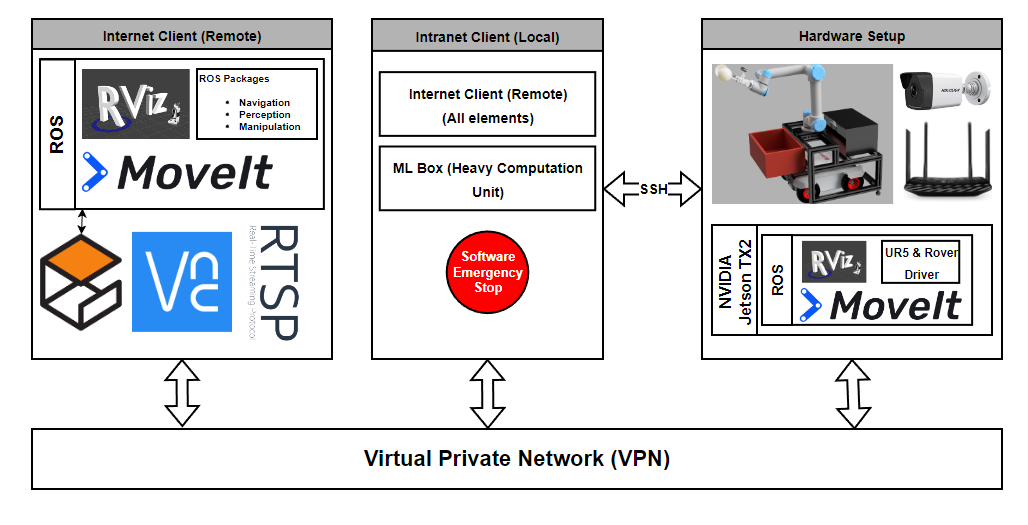}
    \caption{Architecture of the agricultural framework}
    \label{agri:frameowrk}
\end{figure*}

The intranet Client and Hardware setup block are on the same network; hence intranet Client can directly access the hardware by bypassing the VPN. Due to its presence in the same network, there is no latency while communicating with hardware. It consists of a software emergency stop program that instantly stops the hardware. It has more access and permissions to the Hardware Setup block than the Internet Client. It can access the Hardware setup via SSH\cite{ylonen2006secure}. It also streams all the camera feeds of the hardware setup through a VPN. 

Hardware Setup consists of a mobile cobot, IP cameras, and a wireless router for communication. Inside the mobile robot, there is a single board computer board, Nvidia Jetson TX2. It consists of ROS \cite{ros} driver packages for motion control and sensor reading of rover and UR5. It also consists of the MoveIt \cite{inproceedings} ROS package. It can also contain ROS packages for navigation, perception, and manipulation.

Virtual Private Network is used to establish a protected network connection when using public networks. VPN  block is hosted on the cloud due to security reasons and port access restrictions on the Hardware setup local network. The OpenVPN server was used as a VPN server on the cloud. A limited number of authorized clients can access the hardware setup through VPN.  

An Internet client could be a remote developer accessing the Hardware via the Internet through VPN. Once testing on the hardware is done, developers can also migrate the firmware to the hardware setup block. Due to hardware computation limitations, computationally heavy components could be deployed on the Intra client block as it contains more computational resources. 

Figure 4 shows the network diagram of the entire system framework. Client PC in the internet client can get connected to wired or wireless internet to VPN. Scheduled access will be provided to unique client for remote access to Agribot system. As intranet client is located in remote area internet is provided through wireless internet bridge. Wireless bridge transmitter is located where wired internet is accessible. Wireless bridge receiver is located near to greenhouse which is further connected to PoE \cite{mendelson2004all} switch from where various network connections are provided internally. IP cameras, Wifi router and ML box are connected to PoE switch. 4 IP cameras are connected at various location of greenhouse to stream the entire view of greenhouse environment. WiFi router provides wireless connection to Agribot. 

% \begin{figure}
%     \centering
%     \includegraphics[width=\textwidth]{figs/network.png}
%     \caption{Network connection diagram}
%     \label{agri:frameowrk}
% \end{figure}

\begin{figure}[!h]
    \centering
    \includegraphics[width = \columnwidth]{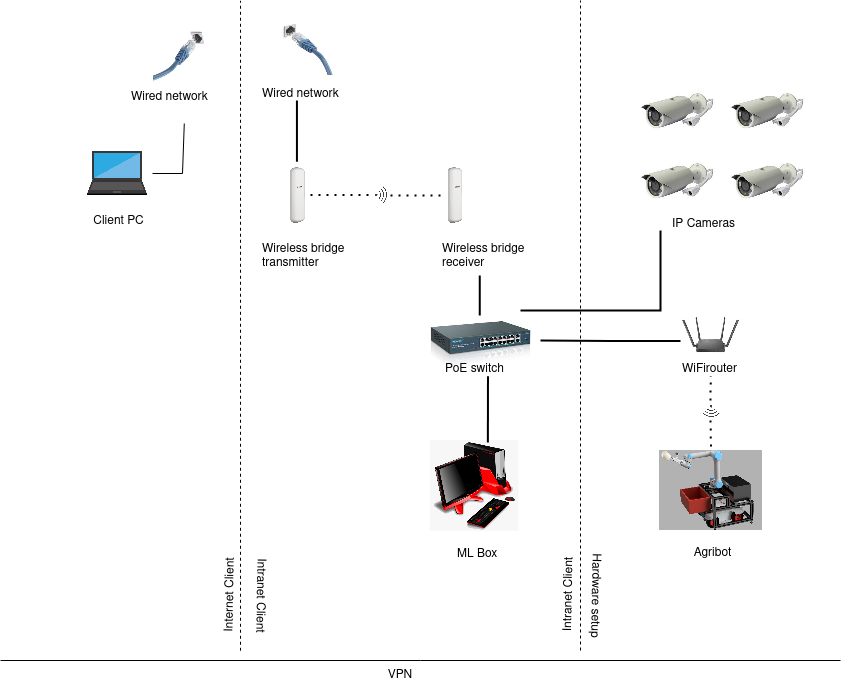}
    \caption{Network connection diagram}
    \label{fig:skid-steer}
\end{figure}

\input{hardware}

\input{software}

%% file: hardware.tex
\subsection{Hardware Stack}

Agribot is designed for the generic autonomous pick-\&-place application, which is generalized as an Autonomous Ground Vehicle (AGV), was previously used in an indoor environment \cite{narvekar2022learning} and with some modification to the design is deployed in a controlled greenhouse condition for the simulation-to-real experiment described in this paper. 

The Agribot consisted of three sub-systems: the mobile base, the arm manipulator, and a trolley-shell

\subsubsection{The Mobile Base}
\begin{figure}[!h]
    \centering
    \includegraphics[width = \columnwidth]{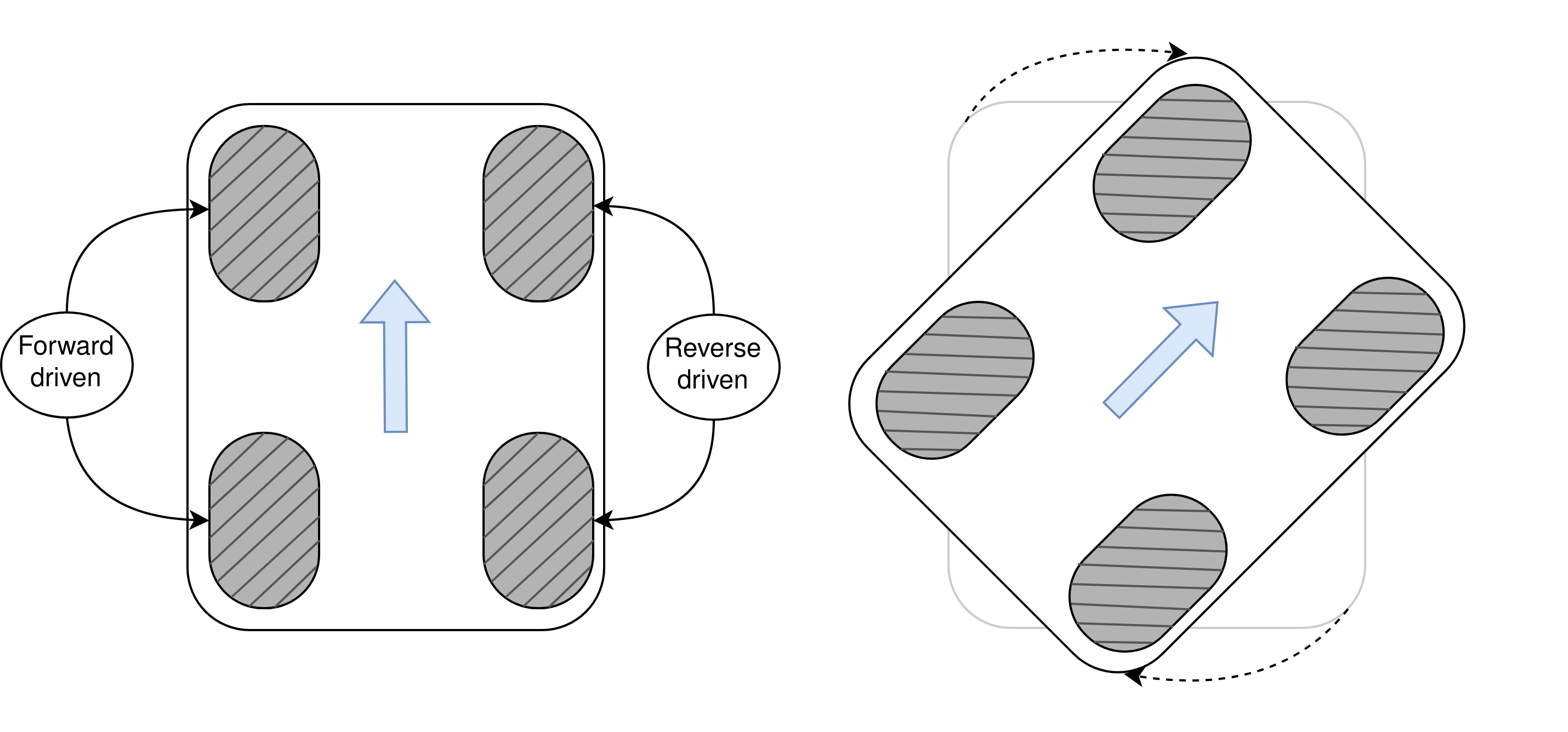}
    \caption{Skid-Steer Drive}
    \label{fig:skid-steer}
\end{figure}

\begin{figure}[!h]
    \centering
    \includegraphics[width = \columnwidth]{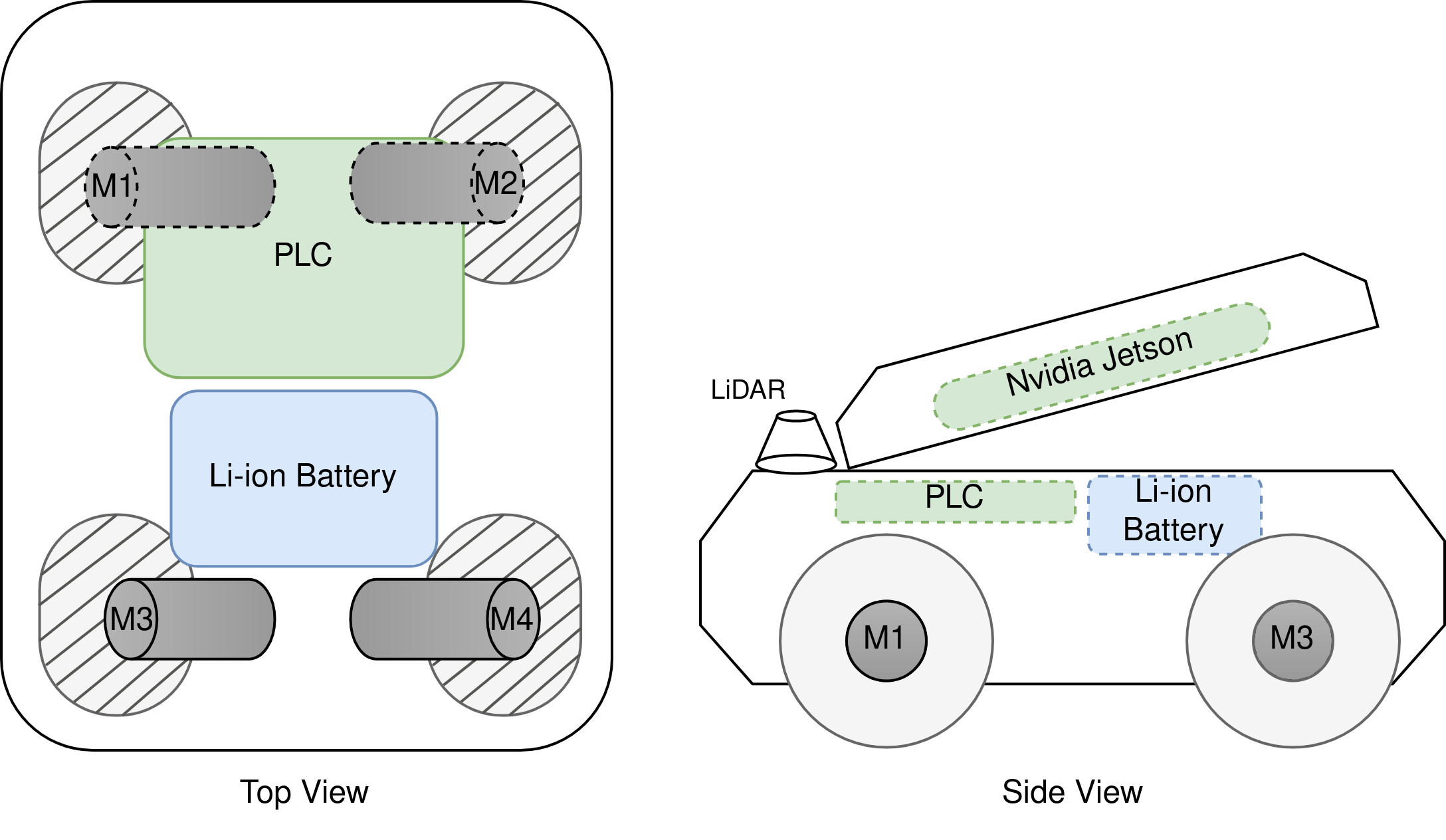}
    \caption{The Mobile Base}
    \label{fig:mobile-base}
\end{figure}

The mobile base uses the skid-steer wheel configuration for locomotion. It works based on the skid/slip caused by the combined differential motion of vertically adjacent wheels driven synchronously. In contrast, the other two wheels rotated in the counter direction for the in-position motion, as seen from the figure-\ref{fig:skid-steer}. The disadvantage of this kind of configuration is difficulty in moving in a straight line due to the inherent nature of 2 similar motors. However, the advantage lies in the simplicity of steering. The skid/slip induces error in the odometer values, compensated using filters and described verbosely in section \ref{sec:software-stack}. The AGV consists of sensors such as SICK 2D LiDAR sensor TiM-P for localization and mapping essential in autonomous navigation; short-range sonar sensors covering the circumference of the body for providing complementary proximity data and one main computing unit, Nvidia’s Jetson TX2 module, working simultaneously with custom PLC circuit board for interfacing motors and power-management units, required to run the entire mobile base, as illustrated in figure-\ref{fig:mobile-base}.

\subsubsection{The Arm Manipulator}
\begin{figure*}[t!]
    \centering
    \includegraphics[width = \textwidth]{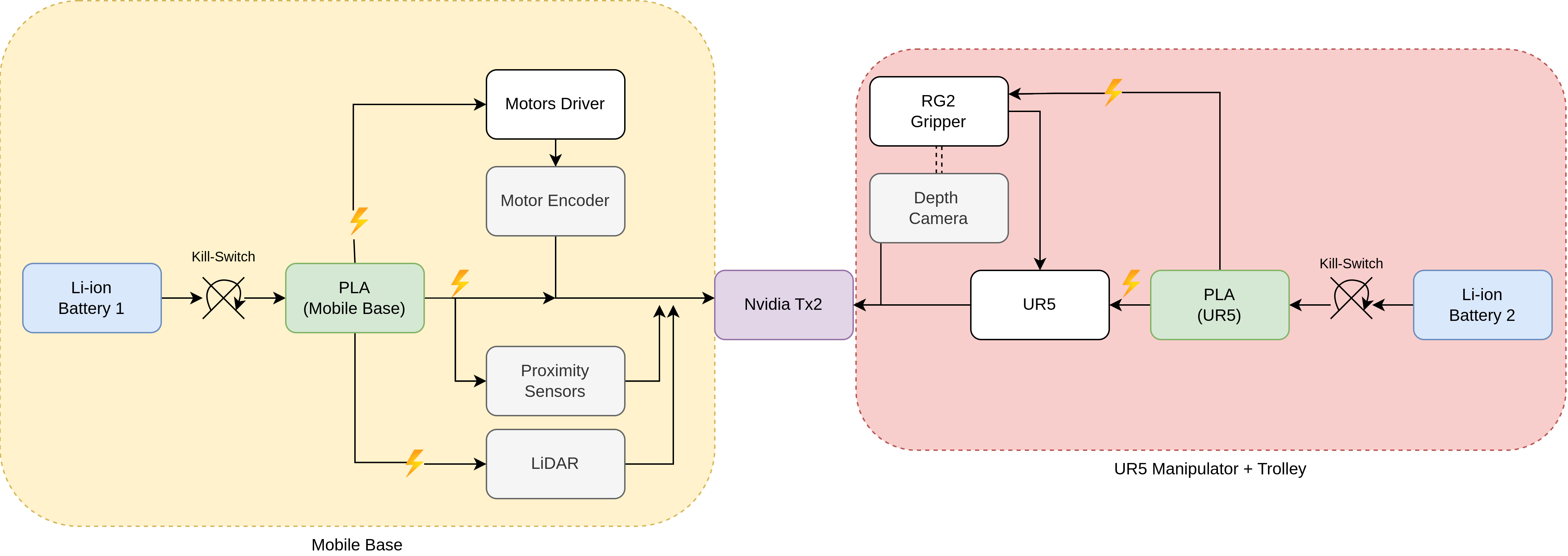}
    \caption{Agribot Over-view}
    \label{fig:agribot-over-view}
\end{figure*}

For pick-\&-place applications, UR5 \cite{ur5_manual}, a robotic arm manipulator, is utilized. The number “5” indicates the payload capacity in kilograms, with a maximum reach of 850mm in reach and a 6-degree of freedom configuration. Driven by default AC supply, it needed various alterations to run on a DC supply instead. Another variant of Universal Robots, UR5e, runs on default DC supply but wasn’t purchased, as the cost of the components required for converting an institutional-donated UR5 from AC-to-DC was much cheaper than buying a new manipulator itself. The challenges with converting the supply type were to match and maintain a constant supply of 20A, 1000W [\cite{ur5_tech_spec}]. A 13C, 4800 mAh Li-ion battery was utilized to ensure a min: 3.7Vx20Ax13C=962W \& max: 4.8Vx20Ax13C=1248W. The overall electrical system can be described in figure \ref{fig:agribot-over-view}. An R2-Gripper [\cite{rojas2016gr2}] is equipped as the end-effector (EE) retrofitted with custom 3D-printed hollow hemispheres, attached at each tip to provide sufficient space for the object of interest, tomatoes while plucking. The EE is also adjunct with a holder for the depth camera, Intel D435i, forming data feedback needed for pose estimation and gripping point assessment, described in section \ref{sec:software-stack} of the desired objects.

\subsubsection{The Trolley-shell}
The third part of the hardware consisted of a trolley, encasing the base and providing scaffolding along with additional room for different aspects of the Agribot, made with 30x30-aluminum rods, providing space for a battery associated with the UR5. A basket was added in the front to place the plucked tomatoes, with the same width as the trolley. The overall system and its Digital twin \cite{jones2020characterising} is illustrated in figure \ref{fig:agribot-sim-to-real}.

% \begin{figure*}[!h]
%     \centering
%     \includegraphics[width =\textwidth]{figs/SIm_HW_comp.png}
%     \caption{Agribot Sim-to-real}
%     \label{fig:agribot-sim-to-real}
% \end{figure*}

%% file: software.tex
\subsection{Software Stack}\label{sec:software-stack}
For the overall pipeline involving the implementation of autonomous navigation, detection, localization, and optimization algorithms, we found that utilizing a state machine to control the execution of these algorithms resulted in system blockages and sudden shutdowns if even one process failed. Additionally, state machines have limited decision-making capabilities, which hinder our ability to perform other tasks simultaneously. Given that computation resources were limited, we also had to consider the potential for "state explosion," which can cause significant problems in system operation. To address these challenges, we opted to parallelize the execution of these algorithms using the Robot Operating System (ROS) \cite{ros}. ROS served as an effective middleware, offering us the tools, libraries, and conventions to simplify the task of creating complex and robust robot behavior. By leveraging the ROS framework, we were able to parallelize the execution of these algorithms, which allowed us to avoid the limitations of state machines. However, parallelization also introduced its own set of challenges, including the demand for additional computation resources. To navigate this, ROS's ability to manage and allocate nodes allowed us to parallelize only a limited set of tasks at a time, optimizing the use of resources while maximizing system performance. Using ROS, we could dynamically reconfigure the system, handle exceptions with more flexibility, and manage a large number of states that the autonomous system can reside in, hence preventing a possible "state explosion." Consequently, we saw an enhancement in the system's robustness and performance.

In our research project, we utilized the perception pipeline (P2Ag) \cite{atar2022p2ag} to enable effective tomato harvesting via instance segmentation and localization techniques. P2Ag was specifically chosen due to its high level of optimization for embedded hardware in terms of performance, computational power, and cost. The pipeline also includes decision-making approaches for harvesting, in addition to perception techniques. Within our implementation, we employed the YOLACT \cite{bolya2019yolact} algorithm for improved efficiency over other segmentation algorithms with comparable performance. By utilizing YOLACT, our agricultural robot was able to locate and approach tomatoes for harvesting or inspection effectively. Overall, the use of P2Ag and YOLACT played a critical role in facilitating efficient and reliable tomato harvesting in our research project.

We recognized that working with real data can be challenging or impossible due to various constraints. Therefore, we opted to leverage simulation techniques as a means of transferring learning from a simulated environment to the real hardware. To accomplish this, we ensured that all settings in the simulation were as close to real-world conditions as possible. We employed the Gazebo simulator to develop, test, and refine our algorithms before deploying them on the physical robot. By doing so, we were able to transfer our algorithms to the real hardware with minimal tuning, except for the learning parameters. This approach enabled us to validate and refine our algorithms in a simulated environment, which ultimately facilitated their successful deployment on the physical robot.

In a dynamic greenhouse environment, classical path planning algorithms that rely on landmarks or occupancy mapping may not perform well due to changes in the environment. Additionally, visual features may not be reliable in such an environment. To address this issue, we developed a novel approach that utilizes marks on each pod and implemented a wall-following algorithm using 2D lidar. This approach allowed our agricultural robot to navigate through the greenhouse environment effectively and efficiently despite changes in the environment. By relying on the marks on each pod and using 2D lidar for wall following, we were able to overcome the limitations of traditional navigation methods and provide a more reliable and accurate solution for greenhouse navigation.

For the robotic manipulator, we employed the Moveit \cite{inproceedings} stack, a popular motion planning framework for robots that runs on the Robot Operating System (ROS). Moveit's inverse kinematics (IK) solvers \cite{6377468} were utilized to enable motion planning for the manipulator. Specifically, we employed a sampling-based Rapidly-exploring Random Tree (RRT) planner in Moveit, which yielded favorable results in terms of motion planning accuracy and efficiency. Our use of Moveit's IK solvers, in conjunction with the RRT planner, enabled us to efficiently plan the robot manipulator's trajectory to accomplish its intended tasks.

\begin{figure*}[h!]
    \centering
    \includegraphics[width = \textwidth]{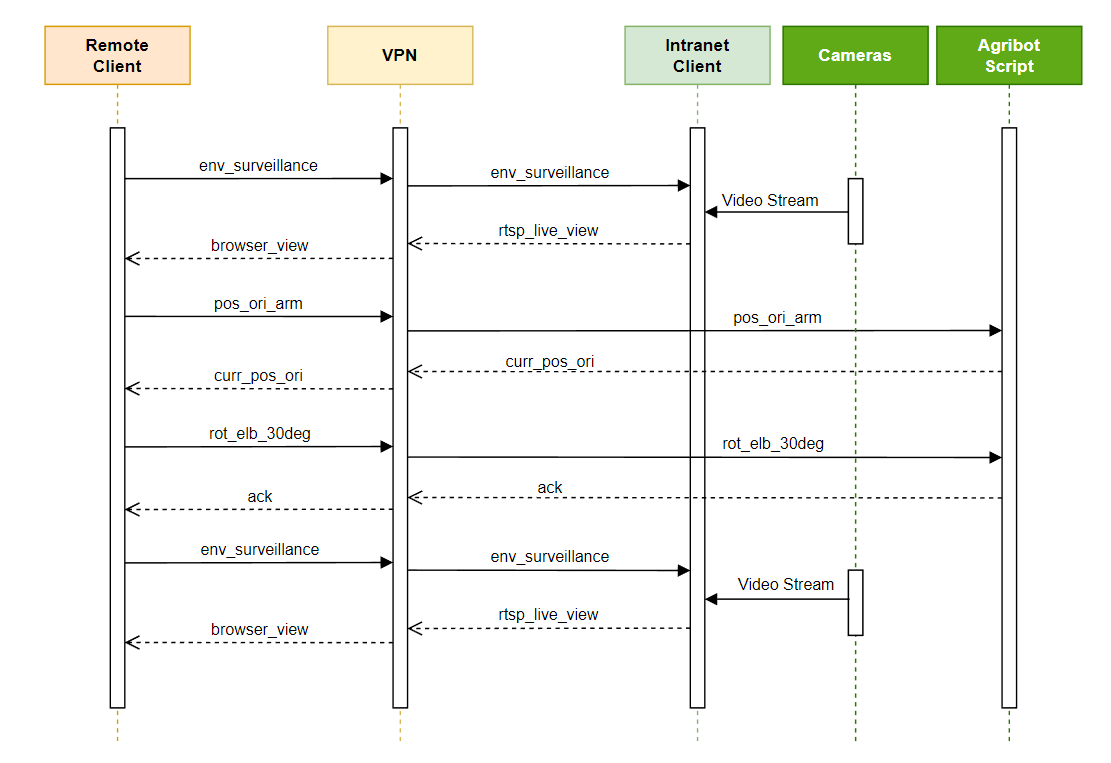}
    \caption{Messages flow diagram}
    \label{agri2:framework}
\end{figure*}

As described in Section 3, Figure \ref{archi} demonstrates a system where a remote client communicates via a secure VPN connection with an intranet client, facilitating data flow integral to the interaction between the robot and surveillance cameras. This data, encompassing the camera view and the positional coordinates of both the robotic arm and the robot itself, is relayed from four surveillance cameras to the intranet client. At the same time, the intranet client receives Robot Operating System (ROS) topics, a feature used for encapsulating inter-process communication. These topics, along with the commands for the robot, which dictate the movements and operations of the robotic arm and mobile robot base, are processed within the intranet client. These commands are first filtered through onboard scripts, which include safety constraints to ensure operations' integrity and prevent harmful actions or movements. After processing this information, the ROS topics and robot commands are published and securely transmitted via the same VPN to the remote client. This setup allows the remote client to effectively control the robot and robotic arm in real time, incorporating an additional layer of interactivity and utility into the surveillance system. This end-to-end secured and seamless connection not only ensures the efficient control and operation of the robot but also leverages real-time data to merge the fields of surveillance and robotics. The onboard scripts add a layer of safety, acting as a crucial barrier that filters out potentially harmful commands, thereby ensuring the safe and smooth operation of the entire system.

Figure \ref{agri2:framework} shows the example of a messages flow diagram to rotate the robotic arm by 30 degrees. As discussed in section 3, agriFrame consists of four blocks: internet client, intranet client, VPN, and Hardware Setup. Internet/remote Client connects through VPN to Intranet Client and hardware. First, to check the status of the robotic arm on the agribot, the remote client needs to check the live video stream of the greenhouse environment. The intranet client receives the message from the remote client for environment surveillance and starts streaming the live video of all four cameras in the greenhouse by an RTSP protocol on a browser. The remote client then needs to check the robotic arm position and orientation. The remote client needs to subscribe to a specific ROS topic to get the present robotic position. The agribot script continuously publishes all the positions and orientations of an agribot, and the remote client gets the position and orientation of the robotic arm. The remote client further sends the command to rotate the elbow of a robotic arm by 30 degrees. The Agribot script running on the Agribot hardware receives the message checks and plans the possible path. It then acknowledges the remote client about the success or failure of the message. Remote clients can see the live robotic arm manipulation on the browser. The connection is lost once the client disconnects through the VPN network, and the AgriFrame can no longer connect.

%% file: results.tex
\section{Results}

The simulator described in section 2 was developed considering all the computation constraints. ROS and Gazebo physics simulator versions are dependent on Ubuntu OS. The AgriFrame simulator requires the system to have Ubuntu 20 operating system with  ROS Noetic version and Gazebo 11 installed on the system. The AgriFrame simulator was also tested in an older version of ROS with the Ubuntu 18 operating system. The performance was much better using ROS noetic and Gazebo 11.

We tested the AgriFrame simulator with different system specifications, as shown in Table 1. System 1 consists of a laptop with 16GB RAM and an external GPU present. System 2 is a desktop computer with 8GB RAM without any external GPU. System 3 is a powerful desktop computer with 16 GB RAM and an external GPU card. Table 2 shows the simulation performance on both systems. The Gazebo simulator only uses GPU for rendering, and all the computation depends on the system processor. Real-time factor (RTF) represents how closely the simulation runs at real speed. If the RTF is significantly below 1, the simulator runs slower than in real-time. The values of RTF range from 0 to 1. From Table 2, we see that the RTF values are dependent on the processor and RAM values. Frames Per Second are increased by very little value in System 3 due to better GPU than in System 1. The current AgriFrame Simulator has a limitation on the number of pluckable tomatoes. Due to the physical calculation constraints of the Gazebo simulator for joints, only a limited number of tomatoes attached to the plants were made pluckable. Other tomatoes are static, meaning they cannot be detached from the plants. We tested with a maximum of 5 pluckable tomatoes in the AgriFrame simulator.

\begin{table}[htbp]
\centering
\begin{tabular}{|l|l|l|l|}
\hline
% \multicolumn{4} \\ \hline
System & Type    & RAM & GPU present \\ \hline
1    & Laptop  & 16  & Yes         \\ \hline
2    & Desktop & 8   & No          \\ \hline
3    & Desktop & 16  & Yes    \\ \hline
\end{tabular}

\caption{ System Specifications}
\label{tab:system-specs}
\end{table}

\begin{table}[htbp]
\centering
\begin{tabular}{|l|l|l|}
\hline
System & RTF & FPS \\ \hline
1 & 0.7 & 50 \\ \hline
2 & 0.6 & 50 \\ \hline
3 & 0.8 & 58 \\ \hline
\end{tabular}

\caption{Simulation Performance}
\label{tab:system-performance}
\end{table}

\begin{figure*}[h!]
    \centering
    \includegraphics[width = 0.9\columnwidth]{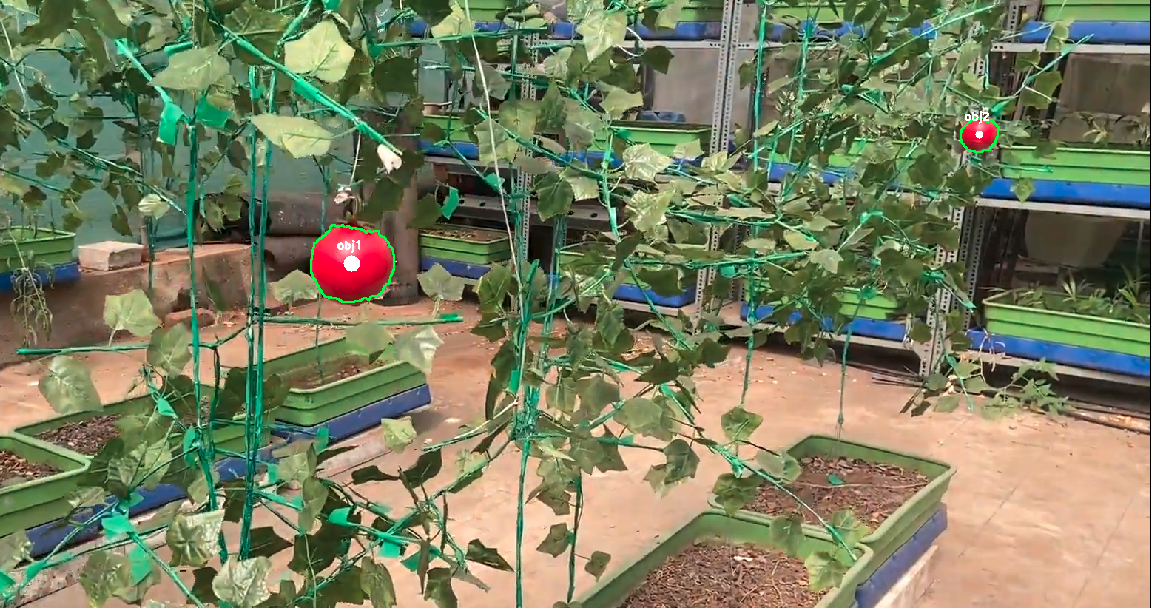}
    \includegraphics[width = 0.9\columnwidth]{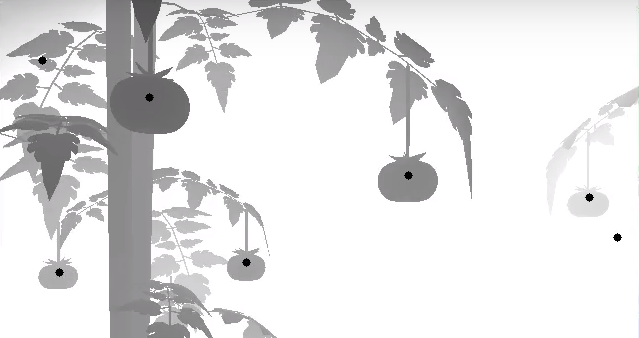}
    \caption{Perception results for real robot and simulated environments}
    \label{perception:real}
\end{figure*}
\label{results}

% \begin{center}
% \begin{tabular}{|l|l|l|l|l|}
% \hline
% \textbf{Model} & \textbf{Backbone} & \textbf{Platform} & \textbf{FPS} \\ \hline
% YOLACT & ResNet50 & Titan Xp & 42.5 \\ \hline
% Mask R-CNN & ResNet101 & Titan Xp & 8.6 \\ \hline
% YOLACT & ResNet50 & Xavier AGX & 8.4 \\ \hline
% YOLACT & ResNet50 & Jetson TX2 & 2.3 \\ \hline
% YOLACT & ResNet101 & GTX 1660Ti & 16\\ \hline
% YOLACT & ResNet50 & GTX 1660Ti & 22 \\ \hline
% \end{tabular}
% \label{tab:compact_analysis}
% \end{center}
% \vspace{-0.25cm}
% \caption{Comparative Analysis of YOLACT and Mask R-CNN}

% \vspace{0.25cm}
\begin{table}[htbp]
\centering
\begin{tabular}{|l|l|l|l|}
\hline
\textbf{Model} & \textbf{Backbone} & \textbf{Platform} & \textbf{FPS} \\ \hline
YOLACT & ResNet50 & Titan Xp & 42.5 \\ \hline
Mask R-CNN & ResNet101 & Titan Xp & 8.6 \\ \hline
YOLACT & ResNet50 & Xavier AGX & 8.4 \\ \hline
YOLACT & ResNet50 & Jetson TX2 & 2.3 \\ \hline
YOLACT & ResNet101 & GTX 1660Ti & 16 \\ \hline
YOLACT & ResNet50 & GTX 1660Ti & 22 \\ \hline
\end{tabular}
\vspace{-0.25cm}
\caption{Comparative Analysis of YOLACT and Mask R-CNN}
\label{tab:compact_analysis}
\vspace{0.25cm}
\end{table}

Table \ref{tab:compact_analysis} provides a comparative analysis of the YOLACT \cite{bolya2019yolact} and Mask R-CNN \cite{maskrcnn} models across different platforms, emphasizing their frames per second (FPS) performance. YOLACT shows remarkable efficiency on the high-end Titan Xp, achieving a top FPS of 42.5, underscoring its optimization for speed on powerful hardware. In contrast, Mask R-CNN, known for its precision in image segmentation, operates at a lower FPS of 8.6 on the same platform, highlighting a trade-off between processing speed and task complexity. Notably, the choice of the backbone network, with YOLACT using ResNet50  \cite{resnet} and Mask R-CNN utilizing ResNet101, plays a crucial role in influencing these performance metrics.

YOLACT's adaptability is further demonstrated on various Nvidia platforms. It achieves a respectable 8.4 FPS on the Nvidia Xavier AGX, which is suitable for embedded systems. At the same time, on the less powerful Nvidia Jetson TX2, the FPS dips to 2.3, indicating the impact of hardware constraints. On the mobile GPU platform Nvidia GeForce GTX 1660Ti, YOLACT maintains a good balance with 22 FPS, showcasing its capability in scenarios that demand both portability and moderate computing power.

Overall, the data underscores the importance of selecting the appropriate hardware for specific image processing needs, balancing between speed, accuracy, and the computational intensity of the task.

% simulaton tested in the competition - computer specs results - curve - cite sahayakbot - 
%participants participated online - cite eyrc 
% best perfroming teams were selected for hardware phase - 5 teams
% Teams were remotely testing  the hardware and running the code - image from ip cam and controling the greenhouse (add image of cam view and map 
The AgriFrame simulator was used in an eYRC-21 competition. The simulator was modeled as per a real greenhouse.  The competition was divided into two parts, i.e., stage 1 and stage 2. Each stage was divided into multiple tasks. In stage 1, "N," no participants used this simulator to test their algorithm. Fig 1 shows the performance of the participants and their computation specifications. We see that our simulator doesn't need very high computation requirements. The top 5 performing teams were selected for stage 2. In stage 2, teams have to work remotely from different regions as shown on the Fig 10  on the agribot hardware in the greenhouse. Green marker on the map in fig 10 shows the geo-location of participants controlling the agribot remotely present in greenhouse with red marker. Teams were provided access to actual hardware with live camera view, as shown in Fig 10, on a slot basis. Each got particular time slot to test their algorithm on the agribot remotely. 

\begin{figure*}[h!]
    \centering
    \includegraphics[width = 1.0\columnwidth]{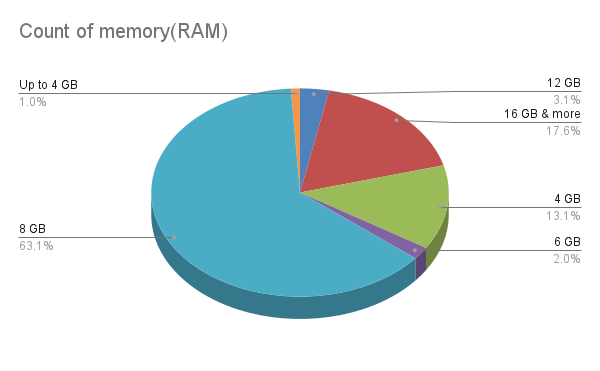}
    \includegraphics[width = 1.0\columnwidth]{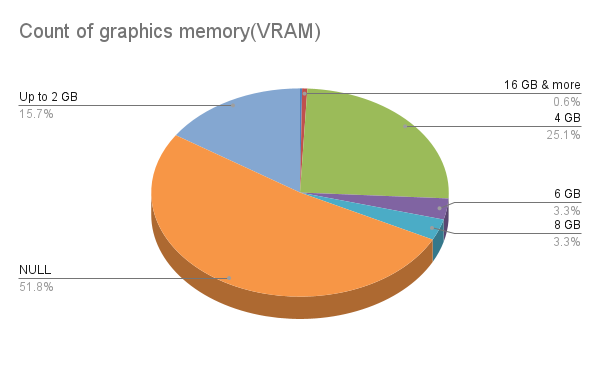}
    \caption{Computer specification of the participants computer}
    \label{perception:real}
\end{figure*}

\begin{figure}[!h]
    \centering
    \includegraphics[width = \columnwidth]{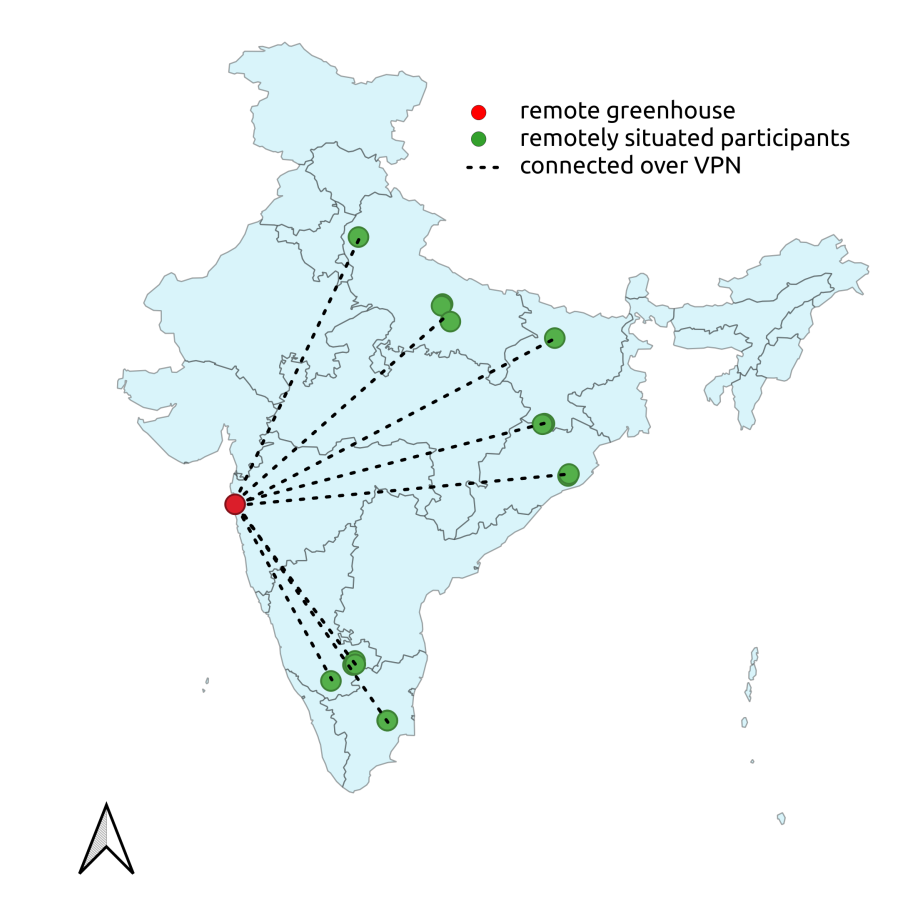}
    \caption{Remote access of hardware}
    \label{fig:skid-steer}
\end{figure}

\begin{figure*}[t!]
    \centering
    \includegraphics[width = \textwidth]{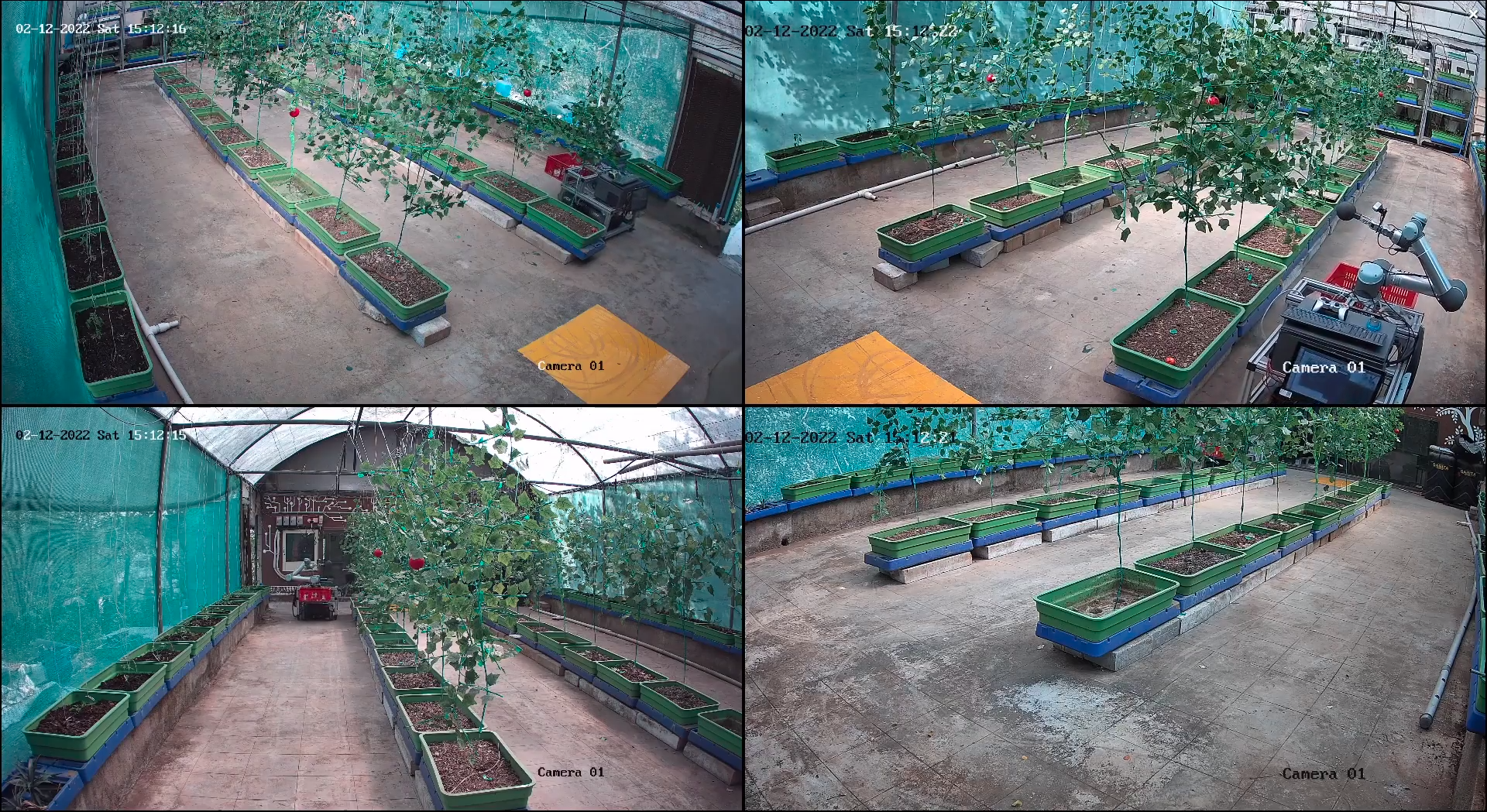}
    \caption{Live camera view from  remote greenhouse}
    \label{fig:agribot-over-view}
\end{figure*}

\begin{figure}[h!]
    \centering
    \includegraphics[width = \columnwidth]{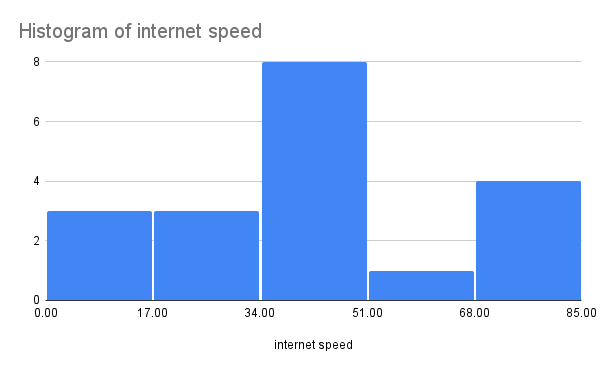}
    \caption{Live camera view from  remote greenhouse}
    \label{fig:agribot-over-view}
\end{figure}

\section{Conclusion and Future scope}
The research covered in this paper demonstrates essential developments in agricultural robotics, from constructing autonomous greenhouse systems to developing frameworks and simulation tools such as agriFrame. These developments provide viable answers to problems, including greenhouse optimization, microclimate variation, and remote control of agricultural equipment. Furthermore, the outcomes showcase how these technologies are feasible and effective on various platforms and how flexible they are concerning varied hardware setups and processing limitations. Embedded technologies such as the Nvidia Jetson TX2 and high-performance GPUs demonstrate adaptability in meeting the various demands of agricultural environments. 

%future scope
%Agrisimulator - limitation of 5 tomatoes for good performance on a computer - optimized by using alternate method 
%ROS 1 was used - disadvantage if ros core is killed on a system - runing on a agribot - remote communication killed - ROS 2 can be used which wont kill the entire system. lag of 2 seconds  can be reduced due to VPN for controlling the bot and camview.
The AgriFrame simulator has a limitation of 5 pluckable tomatoes. This can be optimized by using alternate ways to create pluckable tomatoes in the simulator. ROS 1 was used in the agribot software. One limitation of ROS1 is it needs roscore to run all the processes. Once roscore is killed, the entire process is killed. This can be solved by using ROS2, which is independent of any centralized process like roscore. We found that our AgriFrame hardware access has a lag of 2 sec over the internet.

Realizing the full potential of agricultural robots will need ongoing research and innovation in the sector as we move forward. We can clear the path for a day when agricultural methods will be more sustainable and efficient and fulfill the world's increasing food demand by tackling outstanding issues, improving algorithms, and broadening the practical applications of these technologies. We are at the cusp of a revolution in agriculture and robotics that will transform agricultural cultivation, management, and harvesting practices and provide a more wealthy and sustainable future for future generations. Our architecture can be used in multiple applications, such as industry automation in hazardous environments and space robotics, with some changes in the architecture.
%ucan be used in  industry automation in hazardous environment, space robotics with slight change in architecture

%% file: acknowledgment.tex
\section{Acknowledgement}

The authors wish to express appreciation to the Ministry of Education, Government of India, for their financial support and to the Indian Institute of Technology Bombay for hosting. We would like to extend our sincere thanks to the e-Yantra project team and our esteemed colleagues Ameya Shenoy, Rishikesh Madan, Smit Kesaria, Arjun Sadananda, Simranjeet Singh, Maddu Shravan Murali, and Ajit Harpude for their invaluable contributions to the deliberations and capabilities development. Above all, we take immense pride in e-Yantra’s decade of continuous refinement, enabling an extensive network of students and professionals throughout India with hands-on learning. That has facilitated the fruition of this paper and more.

% rephrase caption : first attenpt of drone shoo